\documentclass[11pt]{article}

\usepackage[preprint]{acl}

\usepackage{times}
\usepackage{latexsym}

\usepackage[T1]{fontenc}

\usepackage[utf8]{inputenc}

\usepackage{microtype}

\usepackage{inconsolata}

\usepackage{graphicx}

\usepackage{amsmath,amsfonts,amssymb}
\DeclareMathAlphabet{\mathbbold}{U}{bbold}{m}{n}
\usepackage{hyperref}
\usepackage[table]{xcolor}
\usepackage{tabularx}
\usepackage{multirow}
\usepackage{makecell}
\usepackage{booktabs}
\usepackage{arydshln}
\usepackage{natbib}

\title{Cross-Tokenizer LLM Distillation through a Byte-Level Interface}

\author{
 \textbf{Avyav Kumar Singh\textsuperscript{1}}\thanks{Equal contribution.}\thanks{Work done during an internship at MediaTek Research. Avyav is now at King's College London, London (United Kingdom).},
 \textbf{Yen-Chen Wu\textsuperscript{1}}\footnotemark[1],
 \textbf{Alexandru Cioba\textsuperscript{1}}\thanks{Work done while at MediaTek Research. Alexandru is now at Orbital Materials, London (United Kingdom).},
 \textbf{Alberto Bernacchia\textsuperscript{1}},
 \textbf{Davide Buffelli\textsuperscript{1}},
\\
 \\
 \textsuperscript{1}MediaTek Research, Cambridge (United Kingdom)
\\
 \small{
   \textbf{Correspondence:} 
   \href{mailto:davide.buffelli@mtkresearch.com}{davide.buffelli@mtkresearch.com}
 }
}

\begin{document}
\maketitle
\begin{abstract}
Cross-tokenizer distillation (CTD), the transfer of knowledge from a teacher to a student language model when the two use different tokenizers, remains a largely unsolved problem. Existing approaches rely on heuristic strategies to align mismatched vocabularies, introducing considerable complexity. In this paper, we propose a simple but effective baseline called \textbf{B}yte-\textbf{L}evel \textbf{D}istillation (BLD) which enables CTD by operating at a common interface across tokenizers: the byte level. In more detail, we convert the teacher's output distribution to byte-level probabilities, attach a lightweight byte-level decoder head to the student, and distill through this shared byte-level interface. Despite its simplicity, BLD performs competitively with---and on several benchmarks surpasses---significantly more sophisticated CTD methods, across a range of distillation tasks with models from 1B to 8B parameters. Our results suggest that the byte level is a natural common ground for cross-tokenizer knowledge transfer, while also highlighting that consistent improvements across all tasks and benchmarks remain elusive, underscoring that CTD is still an open problem.
\end{abstract}

\section{Introduction}
Large Language Models (LLMs) demonstrated unprecedented capabilities in natural language understanding, generation, and reasoning. Their applications are becoming ubiquitous, from conversational agents (e.g., \citep{guo2025deepseek,yang2025qwen3technicalreport,openai2025gptoss120bgptoss20bmodel}) and next-generation search engines \citep{xi2025surveyllmbaseddeepsearch} to tools that assist in scientific discovery \citep{zhang2024comprehensivesurveyscientificlarge} and software development \citep{dong2025surveycodegenerationllmbased}. The remarkable performance of these models, however, is intrinsically linked to their scale with state-of-the-art LLMs often comprising billions of parameters. This size renders their training prohibitively expensive for most research institutes, and often inference becomes prohibitively slow for real-time or on-device applications.

To bridge the gap between the capabilities of large frontier models and the practical constraints of real-world systems, knowledge distillation has emerged as a seminal technique \citep{hinton2015distillingknowledgeneuralnetwork}. Distillation is a process in which a compact \emph{student} model is trained to mimic the behavior of a larger, more powerful \emph{teacher} model. Instead of learning solely from hard labels in a dataset, the student learns from the rich, dense output distribution produced by the teacher. This allows the student to inherit the teacher's sophisticated reasoning patterns while operating with a fraction of the computational footprint. The impact of distillation is already evident across the research environment and the industry, e.g., it enables to speedup the training of small specialized models, and to ``compress'' models and lower costs when serving them at scale \citep{xu2024survey}. 

Despite its success, the standard framework for knowledge distillation is built on a fundamental, yet restrictive, assumption: the teacher and student models must share an identical tokenizer and vocabulary. This is because the most common form of distillation operates at the \emph{logit} level, where the student is trained to match the teacher's probability distribution over a fixed set of vocabulary tokens. If the tokenizers differ, their corresponding vocabularies lead to distinct output spaces. A logit vector of size 50,000 from the teacher cannot be directly compared to a logit vector of size 32,000 from the student. Consequently, performing \emph{cross-tokenizer distillation} (CTD) has been considered infeasible without resorting to approximations or heuristics. These workarounds, such as distilling from generated text samples \citep{kim2016sequence} or attempting to create ad-hoc mappings between vocabularies or hidden states \citep{boizard2025towards,wan2024knowledge,zhang2024dual,Minixhofer2025}, are either computationally inefficient, suffer from significant information loss, or lack a principled theoretical foundation.

The ability to perform principled CTD would unlock powerful new paradigms. First, it would allow us to combine the distinct strengths of diverse models. For instance, one could distill the broad world knowledge of a general-purpose model (e.g., trained with a large, multilingual tokenizer) into a specialized student model equipped with a domain-specific tokenizer optimized for medicine, law, or finance. This would create highly efficient and accurate expert models. Second, it would enable distillation from ensembles of heterogeneous models. For example, training a single student by distilling the collective intelligence of several top-tier open-source models (e.g., DeepSeek \citep{guo2025deepseek}, Qwen \citep{yang2025qwen3technicalreport}, GPT-OSS \citep{openai2025gptoss120bgptoss20bmodel}, etc.), each with its own tokenizer. This would allow the student to learn a consensus of knowledge that potentially surpasses any individual teacher.

In this paper we introduce Byte-Level Distillation (BLD), which sidesteps the vocabulary mismatch in cross-tokenizer distillation by operating at the byte level---a representation shared by all tokenizers. Our method (i) converts the teacher's token-level output distribution to byte-level probabilities using a fast approximation \citep{vieira2025from}, (ii) attaches a lightweight, learnable byte-level decoder head to the student in parallel with its original token-level head, and (iii) performs distillation through this shared byte-level interface. After distillation, the byte-level head is simply removed, leaving a standard token-level model. This approach enables direct and effective knowledge transfer between models with different tokenizers. 

Despite its simplicity, BLD performs competitively with---and on several benchmarks surpasses---substantially more complex CTD methods across tokenizer transfer and cross-model distillation tasks with models ranging from 1B to 8B parameters. At the same time, no method, including ours, achieves consistent gains across all benchmarks, suggesting that CTD remains an open and challenging problem. In summary, our contributions are:
\begin{itemize}
    \item We propose BLD, a simple and alignment-free baseline for CTD that operates through a shared byte-level interface.
    \item We empirically show that this simple approach performs competitively with significantly more complex state-of-the-art CTD methods across a range of tasks.
    \item Through our analysis of the results, we highlight that no existing method---including ours---consistently dominates across benchmarks, and argue that CTD remains a largely open problem deserving further investigation.
\end{itemize}

\section{Related Work}
Our work is positioned at the intersection of three active areas of research: cross-tokenizer knowledge distillation, byte-level language modeling, and methods for converting token-level probability distributions to the byte level.

\paragraph{Cross-Tokenizer Distillation} The challenge of transferring knowledge between models with different tokenizers is a significant hurdle for standard distillation techniques. Several recent works have proposed approximate or heuristic methods to bridge this gap. For instance, some approaches focus on aligning the vocabularies of the teacher and student models through various mapping strategies. \citet{boizard2025towards} introduce a Universal Logit Distillation (ULD) loss based on optimal transport theory, which allows for distillation across different architectures and tokenizers without requiring them to share the same vocabulary. Other works, like \citet{wan2024knowledge} and \citet{zhang2024dual}, explore knowledge fusion and dual-space distillation, respectively, to enable knowledge transfer between heterogeneous models. Similarly, \citet{Minixhofer2025} propose a method for universal cross-tokenizer distillation through approximate likelihood matching. These methods often introduce additional complexity and rely on approximations to align the output spaces of the models. In contrast, our proposed BLD method circumvents this issue by operating at the byte level, a universal interface shared by all tokenizers.

\paragraph{Byte-Level Probability Estimation} A core component of our BLD method is the ability to obtain a byte-level probability distribution from a standard token-based language model. This has been the focus of a number of recent studies. \citet{vieira2025from} present algorithms for converting token-level language models into character-level ones. \citet{phan2025exact} introduce the Byte-Token Representation Lemma, a framework that provides a formal mapping between a model's learned token distribution and its equivalent byte-level distribution. Our work leverages the insights from these works to create a shared byte-level space for distillation.

\paragraph{Byte-Level Language Models} Our work is also related to the growing body of research on byte-level language models, which can be broadly categorized by how they process raw byte sequences.
First are the pure byte-level models, which operate directly on sequences of bytes without any explicit grouping. \citet{xue-etal-2022-byt5}, with their ByT5 model, demonstrated that a standard Transformer architecture can be adapted to process byte sequences effectively, achieving competitive performance with token-level models while being more robust to noise. More recently, \citet{wang2024mambabyte} proposed MambaByte, a token-free model based on the selective state space architecture.
Second are models that use fixed chunking to group bytes into patches. \citet{yu2023megabyte} introduced MEGABYTE, a multi-scale architecture that segments long byte sequences into fixed-size patches, using a local model within patches and a global model across them. \citet{slagle2024spacebyte} proposed SpaceByte, which uses larger Transformer blocks after specific bytes (like spaces) to more efficiently model byte sequences. 
The autoregressive U-Net (AU-Net) of \citet{videau2025bytesideaslanguagemodeling} also falls into this category, as it pools bytes into a multi-scale representation based on fixed rules. 
Third are models that employ learned chunking to dynamically group bytes. Hierarchical Transformers like the Hourglass model from \citet{DBLP:journals/corr/abs-2110-13711} and the dynamic pooling mechanism from \citet{nawrot-etal-2023-efficient} laid the groundwork for more flexible byte-level processing. More recent works have built on this, such as the Byte Latent Transformer (BLT) from \citet{pagnoni-etal-2025-byte}, which encodes bytes into dynamically sized patches based on next-byte entropy, and MrT5 from \citet{kallini2025mrt}, which uses dynamic token merging. The H-Net model from \cite{hnet} takes this a step further with a dynamic chunking mechanism that learns content- and context-dependent segmentation directly from the data, effectively creating an end-to-end, tokenizer-free model. While our method does not involve using byte-level models, it can be used to distill information from token based model into byte-level ones.

\section{Our Method}

\subsection{Preliminaries}
\label{sec:preliminaries}
Let $\Sigma$ be the alphabet containing all bytes, i.e., $\{1, 2, \dots, 256\}$, and let $\Sigma^{*}$ be the set of all sequences over the alphabet.
Given a vocabulary $V \subseteq \Sigma^{*}$, which determines all the possible tokens, a tokenizer is a deterministic function that maps sequences of bytes to sequences of tokens: $\mathcal{T}:\Sigma^{*} \rightarrow V^{*}$, where $V^{*}$ indicates the set of all sequences composed of tokens from the vocabulary $V$. We also define a decoder function $\mathcal{D}:V^* \rightarrow \Sigma^*$ as the function that ``maps back'' from a sequence of tokens to a sequence of bytes. We can assume that the decoder function is the inverse of the tokenizer, i.e., $\mathcal{D}(\mathcal{T}(\{b_1, b_2, \dots, b_{N_b} \})) = \{b_1, b_2, \dots, b_{N_b} \}$, with $N_b$ indicating the length of the byte sequence, though this is not always the case in practice\footnote{This is because in practice tokenizers involve some pre-tokenization steps which are not reversible, like for example normalizing Unicode characters.}. 

When performing distillation, the goal is to transfer knowledge from a teacher model to a student model.  
The teacher model has an associated vocabulary $V_T$, tokenizer $\mathcal{T}_{T}$, and decoder $\mathcal{D}_T$. The teacher model can be seen as a function mapping a given tokenized input sequence into a probability distribution over its vocabulary indicating the probability of the next token, $f_T: \mathcal{T}_T(\Sigma^{*}_{T}) \rightarrow \Delta(V_T)$, where $\Delta(V_T)$ is the probability simplex over the vocabulary. 
Similarly, the student model also has a vocabulary $V_S$, tokenizer $\mathcal{T}_{S}$, and decoder $\mathcal{D}_S$, which may differ from those of the teacher.

In standard distillation approaches, given a dataset of tokenized sequences $\mathcal{Z} = \{ s_1, s_2, \dots \}$, each one composed of multiple tokens $s_i = \{t_1, t_2, \dots, t_{\lvert s_i \lvert} \}$, the student model parameters are updated by minimizing the following loss function
\begin{align}
    \mathcal{L} = \sum_{s_i \in \mathcal{Z}}
     \frac{1}{\lvert s_i \lvert}
    & \biggl(
    \sum_{t_j \in s_i} \text{CE}(\delta(t_j), f_S(t_{< j})) + \nonumber \\ 
    & \text{KL}(f_T(t_{< j}), f_S(t_{< j}))
    \biggr)
    \label{eq:loss}
\end{align}
where $\delta(t_j)$ is the delta function which is zero everywhere except at the index of token $t_j$ for which it is equal to 1, $t_{< j}$ indicates the sequence of tokens up to the $j$-th token excluded, $\text{CE}$ indicates cross-entropy, and $\text{KL}$ indicates the Kullback–Leibler divergence.
The first term in equation \ref{eq:loss}, the cross entropy, is the standard next token prediction loss, while the second term, the KL divergence, is responsible for transferring knowledge from the teacher to the student. Notice however that for the latter to be well defined, it requires teacher and student to have the same vocabulary, which in practice usually leads to sharing also the same tokenizer, although in theory the it could be different between the two.
Recently, several works have introduced heuristic or approximate strategies to overcome this issue \citep{boizard2025towards,wan2024knowledge,zhang2024dual,Minixhofer2025}. These approaches  require identifying some form of alignment between tokenizations and introducing additional heuristic losses.
Our approach instead overcomes these challenges by performing distillation at the byte level.

\paragraph{From BPE-level to Byte-Level Probabilities.} Given a sequence of bytes $\{b_1, b_2, \dots, b_{N_b} \}$ and a teacher model $f_T$ with vocabulary $V_T$ and tokenizer $\mathcal{T}_T$, \citet{phan2025exact} and \citet{vieira2025from} show that it is possible compute the probability of generating a sequence of bytes using the model $f_T$ by summing the probabilities that the model assigns to all the \emph{coverings} of the byte sequence. 
Let us define a \emph{covering}, associated to the teacher model, for a byte sequence $\{b_1, b_2, \dots, b_{N_b} \}$ as the set containing all the sequences of tokens that ``cover'' the sequence of bytes when decoded, i.e., 
\begin{align}
    \text{cover}_T&(b = \{b_1, b_2, \dots, b_{N_b} \}) =  \nonumber  \\
    \{
    &\{t_1, t_2, \dots, t_m \} \in V^*_{T} \quad | \quad \exists i \in \mathbb{Z}^{>0} \text{ s.t. }  \nonumber \\
    &\mathcal{D}_T(\{t_1, t_2, \dots, t_{m-1}\}) = b_{< i} 
    \text{ and } \nonumber \\
    & b_{\ge i} \text{ is a prefix of } \mathcal{D}(t_{m}) \}
    \}
\end{align}
We can now compute the probability assigned by the teacher to a byte sequence $b=\{b_1, b_2, \dots, b_{N_b} \}$ as
\begin{equation}
    P_T(b) = \nonumber 
    \sum_{y_i \in \text{cover}_T(b)} \prod_{t^{(i)}_j \in s_i} f_T\left( t^{(i)}_j \lvert t^{(i)}_{<j} \right)
   \label{eq:byteprob}
\end{equation}

From this we can straightforwardly obtain the conditional probabilities for each single byte in the sequence as 
\begin{equation}
    P_T(b_i \lvert b_{<i}) = \frac{P_T(\{b_1, b_2, \dots, b_{i}\})}{P_T(\{b_1, b_2, \dots, b_{i-1}\})}
    \label{eq_marginal}
\end{equation}
The above procedure can be quite expensive computationally, but \citet{vieira2025from} provide a fast approximation, which we use for our method. More details are provided in Appendix \ref{app:approx_settings}.

\paragraph{A naive approach to byte level CTD.} Given that we can extract the probabilities at the byte level from any token based model, one might think of ``going back'' from byte level to a different token level to perform CTD. In fact, a naive approach for byte-level CTD, once the probabilities $P_T(b_i \lvert b_{<i})$ at the byte level are extracted from the teacher for a given sequence, could be to use them to construct the probabilities of a tokenized version of the sequence in which the student's tokenizer is used instead. 
In more detail, given a sequence $b = \{b_1, b_2, \dots\}$, we can tokenize it using the student's tokenizer into a sequence of tokens $\{y_1, y_2, \dots \} = \mathcal{T}_S(b)$, and then compute the probability of each possible token (as this is needed for the KL term in the distillation loss) in $V_S$ as follows

\begin{align}
    \forall t &= \{b^{(t)}_{1}, \dots, b^{(t)}_{k}\} \in V_S,  \nonumber \\   
    & P(y_i = t \lvert y_{< i}) = \prod_{b^{(t)}_j \in t} P_T \left( b^{(t)}_j \lvert b^{(t)}_{<j}, y_{<i} \right) \label{eq:naive}
\end{align}
where, with a slight abuse of notation, we use $P_T \left( b^{(t)}_j \lvert b^{(t)}_{<j}, y_{<i} \right)$ to indicate the probability assigned by the teacher to the $j$-th byte of token $t$ given all previous bytes in the whole sequence. This quantity is computed using the equations presented above.
The advantage of this approach is that there is no need to add any module to the original architecture of the student (which instead is required in our method). On the other side, this approach has several issues that make it impractical. First, equation \eqref{eq:naive} requires the computation of $\lvert V_S \lvert$ probabilities -- which in practice is between 30000 and 250000 -- for each token in the sequence (where the sequence is tokenized according to the student's tokenizer $\mathcal{T}_S$), which would be computationally prohibitive. Second, if the byte level probabilities are computed with an approximate method, the errors will compound when computing equation \eqref{eq:naive}.

\subsection{Byte-Level Interface for Distillation}
\begin{figure*}
    \centering
    \includegraphics[width=0.99\linewidth]{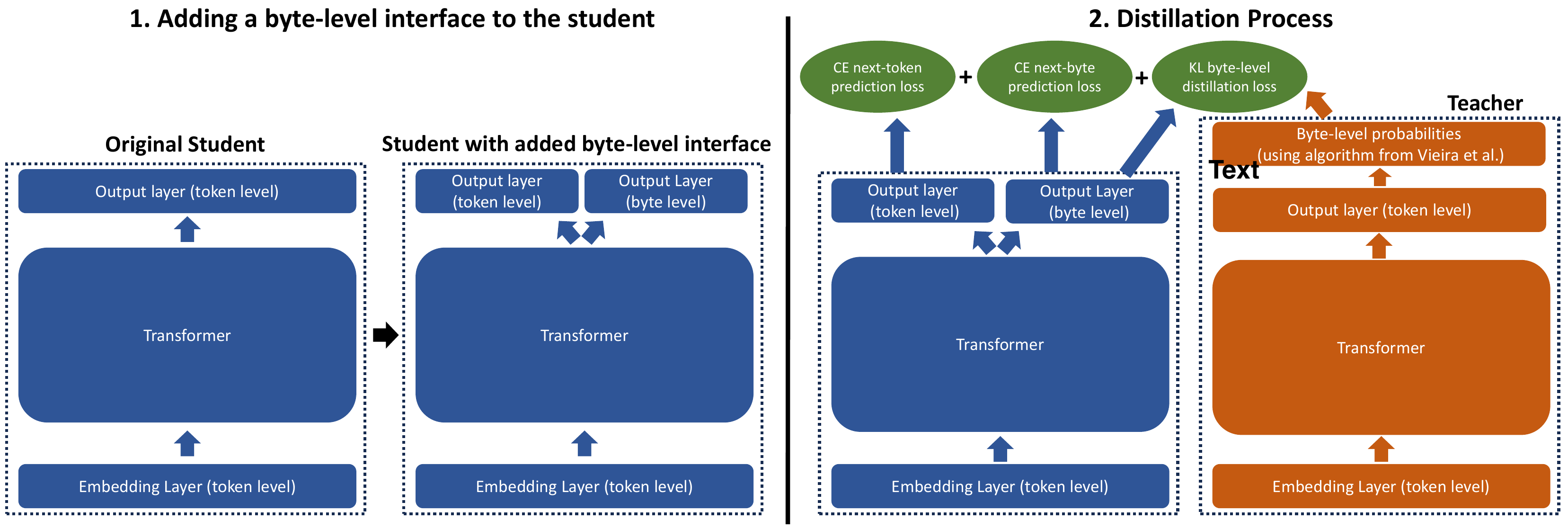}
    \caption{Representation of our Byte-Level Distillation (BLD) method composed of two steps. Step 1 adds a byte-level interface to the student model. Step 2 performs distillation by transferring knowledge from the teacher to the student using the shared byte-level interface. Additional next-token prediction and next-byte prediction losses are also used following standard distillation approaches. The byte-level interface can be removed at the end of the process.}
    \label{fig:our_method}
\end{figure*}

Our method, called Byte Level Distillation (BLD), can be divided into two steps which we present below.
A schematization of BLD can be found in Figure \ref{fig:our_method}.

\paragraph{Step 1: byte-level interface.} The first step is to enable teacher and student models to share knowledge through the byte level. For the teacher we can use the approach presented in Section \ref{sec:preliminaries} to compute byte-level probabilities, but for enabling training of the student we need to introduce a new module to it. We start from a pretrained student model. The model is composed of a tokenizer $\mathcal{T}_S: \Sigma^{*} \rightarrow V^{*}_S$ with a respective decoder $\mathcal{D}_S:V^*_S \rightarrow \Sigma^*$, an encoder $E:V^{*}_S \rightarrow \mathbb{R}^{N \times d}$ (typically a learnable embedding matrix with one row for each element of the vocabulary $V_S$), a transformer $H:\mathbb{R}^{N \times d} \rightarrow \mathbb{R}^{N \times d}$, and an output layer $O: \mathbb{R}^{N \times d} \rightarrow \mathbb{R}^{N \times \lvert V_S \lvert }$. Here $N$ is the input sequence length (in terms of numbers of tokens from the vocabulary $V_S$), and $d$ is the dimension of token embeddings and hidden representations (we assume they are the same for simplicity of presentation but in practice hidden dimensions at every layer of the transformer can be different from the dimensions of token embeddings). We now add a new learnable module to the student model. In more detail, in parallel to the existing token-level decoder $O$, we a add byte-level decoder: $O_b: \mathbb{R}^{N \times d} \rightarrow \mathbb{R}^{ N_b \times \lvert \Sigma \lvert}$, where $N_b$ is the length of the input sequence in terms of bytes. With this we have effectively added a \emph{byte-level interface}  to the output of the student model\footnote{The byte-level decoder can be pre-trained while keeping the rest of the weights fixed for additional stability, but in our experiments we found that it is not necessary.}.
    
\paragraph{Step 2: distillation.} Given a teacher model, we use the method of \citet{vieira2025from} to obtain $P_T(b_i \lvert b_{<i})$ for each sequence $x_i = \{b_1, b_2, \dots\}$ in a given dataset $\mathcal{D}$. We can now perform distillation without requiring any specific alignment or heuristic as we have the probabilities at the byte-level obtained from the teacher model, and our student model has an output interface at the byte level. During distillation the loss is a combination of next-byte cross entropy loss, KL divergence at the byte level, and next-token cross entropy loss\footnote{The next token cross entropy loss is added to ensure the weights of the token-level decoder $O$ get updated.}. Formally, let $f_S: \mathcal{T}_S(V^{*}_{S}) \rightarrow \Delta(V_S)$ be the function at the token level for the student model obtained by composing $f_S(t) = O(H(E(t)))$ and let $f^{(b)}_S: \mathcal{T}_S(V^{*}_{S}) \times \mathbb{Z}^{>0} \rightarrow \Delta(\Sigma)$ be the function with the byte-level interface for the student model, i.e.,  $f^{(b)}_S(t, j) = O_b(H(E(t))) [j]$ (where ``$[j]$'' indicates selecting the $j$-th byte of the output), then the full loss for distillation is:
    \begin{align}
        \mathcal{L} =& \sum_{\substack {x_i \in \mathcal{Z}, \\ \{t_1, t_2, \dots, t_k\}=\mathcal{T}_S(x_i), \\ t_i= \{b^{(i)}_1, \dots, b^{(i)}_{n_{i}} \} } }
        \frac{1}{k}
        \sum_{\ell=1}^{k}
        \biggl[
         \text{CE}\left( \delta(t_{\ell}), f_s(t_{< \ell}) \right)
        \nonumber \\
        & + \frac{1}{n_{\ell}}
        \sum_{j = 1}^{n_{\ell}}
        \text{CE}\left( \delta(b^{(\ell)}_j), f^{(b)}_S(t_{< \ell}, j) \right) + \nonumber \\
        & \text{KL}\left( P_T(b^{(\ell)}_j \lvert b^{(\ell)}_{< j}, t_{< \ell}), f^{(b)}_S(t_{< \ell}, j) \right)
        \biggr]
        \label{eq:our_loss}
    \end{align}
where $P_T(b^{(\ell)}_j \lvert b^{(\ell)}_{< j}, t_{< \ell})$ indicates the probability assigned by the teacher to the $j$-th byte in the $\ell$-th token given all bytes in the sequence (including those from  tokens prior to the $\ell$-th) up to the $(j-1)$-th byte of the $\ell$-th token.
All or a subset of the parameters of the model can be updated during distillation, except from the byte-level output layer which must be updated if not pre-trained first.

After distillation, we  remove the byte-level interface $O_b$, and thus keeping only the token-level output layer $O$. It is also possible to instead keep the byte level output layer if one is interested in generating outputs in terms of bytes or combinations of tokens and bytes. 
In our experiments, as byte-level decoder $O_b$, we use a simple linear projection for $O_b$ with $N_b$ fixed to 10, which means that for tokens that span more than 10 bytes, supervision signal will be provided only for the first ten. We validate our choice experimentally as shown in Appendix \ref{app:byte_level_head}. A different approach could be to have a small autoregressive layer to accomodate different values of $N_b$.
We leave these directions for future work.

\newcolumntype{R}{>{\raggedleft\arraybackslash}X}
\begin{table*}[ht]
\centering
\small
\setlength\tabcolsep{2pt}
\renewcommand{\arraystretch}{1.0}
\begin{tabularx}{\linewidth}{llRRRRRRR}
\toprule
\multirow{2}{*}{\textbf{Model}} & \multirow{2}{*}{\textbf{Method}} & \multicolumn{7}{c}{\textbf{Benchmark}} \\
& & {\scriptsize PiQA} & {\scriptsize ARC-C} & {\scriptsize BoolQ} & {\scriptsize MMLU} & {\scriptsize AGI-EN} & {\scriptsize AGI-ZH} & {\scriptsize IFEval}\\
\midrule\noalign{\vskip -0.6ex}
\rowcolor{gray!20} \multicolumn{2}{c}{\textit{original (Llama3.2 3B IT)}} & 75.46 & 45.73 & 78.41 & 60.50 & 35.27 & 42.93 & 66.31 \\
\noalign{\vskip -0.4ex}\midrule
\multirow{5}{*}{$\rightarrow$ Qwen2} & SFT & 74.54 & 41.89 & 76.48 & 57.11 & 30.47 & 34.30 & 26.74 \\
 & DSKD & 62.95 & 28.84 & 71.80 & 50.48 & 26.12 & 34.18 & 28.13 \\
 & MinED & 75.35 & 42.58 & 78.65 & 58.20 & 34.68 & 34.76 & \textbf{62.83} \\
 & ALM + SFT & 75.46 & \textbf{45.82} & \textbf{79.36} & \textbf{58.86} & \textbf{36.64} & 35.27 & 58.51 \\
\cdashline{2-9}\noalign{\vskip 0.5ex}
 & BLCTD (Ours) & \textbf{75.68} & 43.26 & 77.34 & 58.29 & 31.98 & \textbf{35.97} & 30.58 \\
\bottomrule
\end{tabularx}
\caption{Results of transferring Llama3.2 3B \citep{grattafiori2024llama3herdmodels} to the Qwen2 tokenizer \citep{yang2024qwen2technicalreport}. \textit{original} denotes the original model without transfer. \textit{ARC-C} refers to Arc-Challenge. \textit{AGI-EN} and \textit{AGI-ZH} refer to the English and Chinese splits of AGIEval.}
\label{table:transfer_qwen}
\vspace{-0.3cm}
\end{table*}

\begin{table*}[ht]
\centering
\small
\setlength\tabcolsep{2pt}
\renewcommand{\arraystretch}{1.0}
\begin{tabularx}{\linewidth}{llRRRRRRR}
\toprule
\multirow{2}{*}{\textbf{Model}} & \multirow{2}{*}{\textbf{Method}} & \multicolumn{7}{c}{\textbf{Benchmark}} \\
& & {\scriptsize PiQA} & {\scriptsize ARC-C} & {\scriptsize BoolQ} & {\scriptsize MMLU} & {\scriptsize AGI-EN} & {\scriptsize AGI-ZH} & {\scriptsize IFEval}\\
\midrule\noalign{\vskip -0.6ex}
\rowcolor{gray!20} \multicolumn{2}{c}{\textit{original (Llama3.2 3B IT)}} & 75.46 & 45.73 & 78.41 & 60.50 & 35.27 & 42.93 & 66.31 \\
\noalign{\vskip -0.4ex}\midrule
\multirow{5}{*}{$\rightarrow$ Byte} & SFT & 67.30 & 31.57 & \textbf{73.00} & 38.95 & 26.05 & 35.18 & 24.70 \\
 & DSKD & 64.47 & 31.31 & 60.34 & 37.62 & 23.74 & 33.36 & 23.98 \\
 & MinED & 67.41 & \textbf{32.94} & 65.32 & \textbf{39.84} & 27.52 & 33.90 & 31.89 \\
 & ALM + SFT & 66.32 & 31.57 & 71.41 & 39.15 & \textbf{27.66} & \textbf{35.39} & 29.74 \\
\cdashline{2-9}\noalign{\vskip 0.5ex}
 & BLCTD (Ours) & \textbf{67.52} & 30.89 & 69.85 & 39.06 & 26.44 & 34.57 & 25.43 \\
\bottomrule
\end{tabularx}
\caption{Results of transferring Llama3.2 3B \citep{grattafiori2024llama3herdmodels} to byte-level tokenization. \textit{original} denotes the original model without transfer. \textit{ARC-C} refers to Arc-Challenge. \textit{AGI-EN} and \textit{AGI-ZH} refer to the English and Chinese splits of AGIEval.}
\label{table:transfer_byte}
\vspace{-0.3cm}
\end{table*}

\section{Experiments}
To evaluate our approach, we follow the experimental procedure of \citet{Minixhofer2025} which considers three tasks: tokenizer transfer across different BPE tokenizers, tokenizer transfer from BPE to byte, and cross-tokenizer distillation.

\paragraph{Training setup.}
We fine-tune the student backbone with LoRA \citep{hu2022lora}, applying rank $r=64$ updates to the query and value projection matrices while keeping all other backbone weights frozen. For tokenizer transfer experiments, the embedding matrix and LM head are re-initialised using Fast Vocabulary Transfer (FVT) \citep{gee-etal-2022-fast}: tokens present in both vocabularies are initialised by directly copying the corresponding source embedding; tokens absent from the source vocabulary are initialised as the mean of their constituent sub-token embeddings, falling back to a random Gaussian sample drawn from the source
embedding distribution when no decomposition is available. The byte-level decoder head $O_b$ is a lightweight module consisting of 10 parallel linear projections from the model's hidden dimension to the byte vocabulary (260 tokens representing the 256 bytes and 4 special tokens for: beginning of sequence, end of sequence, padding, and out-of-vocabulary), enabling each token position to predict up to 10 bytes simultaneously (see Appendix \ref{app:byte_level_head} for a validation of this approach). We optimize with AdamW \citep{loshchilov2018decoupled} using a cosine learning rate schedule with linear warm-up. Full hyperparameter details are provided in Appendix~\ref{app:hyperparameters}. Importantly, we use the same SFT backbone for all considered distillation methods.

\paragraph{Training datasets} For the BPE tokenizer transfer and byte tokenizer transfer experiments, we train on the Tulu-3 SFT mixture \citep{lambert2024tulu3}. Byte-level teacher probabilities are pre-computed offline for this dataset using the fast approximation of \citet{vieira2025from}, as described in Appendix~\ref{app:approx_settings}. For the cross-tokenizer distillation experiment (OpenMath2-Llama3.1-8B $\rightarrow$ Gemma2 2B), we train on the OpenMathInstruct-2 dataset \citep{toshniwal2024openmathinstruct}. 

\paragraph{Validation datasets}
For the tokenizer transfer experiments, we use the no-robots split of Tulu-3 \citep{rajani2023norobots} as a held-out validation set; this subset spans a diverse range of tasks---including coding, mathematics, and general reasoning---making it a representative signal for general-purpose capability. For the cross-tokenizer distillation experiment, we randomly sample approximately 1,000 examples from OpenMathInstruct-2 as a held-out validation set.

\subsection{BPE Tokenizer Transfer}
We first evaluate our method on the task of \emph{tokenizer transfer} between two different BPE tokenizers. This involves selecting a pre-trained model, in our case LLama 3.2 3B \citep{grattafiori2024llama3herdmodels}  and replacing its tokenizer with the BPE tokenizer from Qwen 2 \citep{yang2024qwen2technicalreport}. The procedure involves replacing embedding and output projection layers with uninitialized layers in accordance (in terms of dimensionalities) with the new tokenizer, and distilling from the original model to the modified one. We present results in Table \ref{table:transfer_qwen}.

Table~\ref{table:transfer_qwen} shows that BLD performs competitively but does not uniformly dominate. It achieves the highest scores on PiQA (75.68) and AGI-ZH (35.97), and recovers performance close to the original model on PiQA, MMLU, and BoolQ, demonstrating that distillation through the byte-level interface successfully transfers general knowledge after tokenizer replacement. ALM~+~SFT is the strongest overall competitor, leading on four of seven benchmarks (ARC-C, BoolQ, MMLU, AGI-EN). The most notable weakness of BLD is instruction following: its IFEval score (30.58) lags far behind MinED (62.83) and ALM~+~SFT (58.51), both of which retain near-original IFEval performance. This suggests that the byte-level distillation objective does not sufficiently preserve the structured output behaviour required for instruction following. DSKD performs worst across all benchmarks, confirming that direct distribution alignment without vocabulary alignment is ineffective in this setting.

\subsection{BPE-to-byte Tokenizer Transfer}
We now repeat the same \emph{tokenizer transfer} task as the previous section, but this time we transfer from a BPE tokenizer to byte-level. This can be seen as adapting LLama 3.2 3B \citep{grattafiori2024llama3herdmodels} to be a byte-level model. The procedure involves replacing embedding and output projection layers with uninitialized layers compatible with a byte-level tokenizer, and distilling from the original model to the modified one. We present results in Table \ref{table:transfer_byte}.

Results in Table~\ref{table:transfer_byte} show that transferring to byte-level tokenization is substantially harder than BPE-to-BPE transfer: all methods suffer large degradations across every benchmark (e.g., MMLU drops approximately 21 points and ARC-C approximately 13 points relative to the original model), reflecting the challenge of adapting a model trained on subword tokens to a much finer-grained representation. In this setting, BLD ranks first on PiQA (67.52), though the margin over MinED (67.41) is negligible. Performance leadership is fragmented across methods: SFT leads on BoolQ (73.00), MinED on ARC-C (32.94) and MMLU (39.84), and ALM~+~SFT on AGI-EN (27.66) and AGI-ZH (35.39). The spread between methods is noticeably narrower than in Table~\ref{table:transfer_qwen}, suggesting that in this harder regime all approaches converge to a similar performance ceiling. DSKD again performs worst across most benchmarks. Overall, no method establishes a clear advantage, and the collective degradation relative to the original underscores that byte-level tokenizer transfer remains an unsolved challenge. 

\subsection{Cross-Tokenizer Distillation}
Finally, we perform CTD across different models with different tokenizers. In more detail, we distill the maths-specialised OpenMath2-Llama3.1-8B \citep{toshniwal2024openmathinstruct} into Gemma2 2B \citep{gemmateam2024gemma2improvingopen}.
Results are shown in Table \ref{table:cross_tokenizer_distil}. 

Table~\ref{table:cross_tokenizer_distil} shows that BLD achieves the highest GSM8K score (62.55), modestly outperforming ALM~+~SFT (61.56) and SFT (59.29), and represents a meaningful gain over the uninitialised Gemma2 2B IT baseline (51.48). However, SFT leads on MATH (22.40 vs.\ 20.08 for BLD), suggesting that BLD's advantage over SFT is task-dependent and does not generalise uniformly across mathematical reasoning benchmarks. 
Despite BLD's result, the gap to the teacher (87.26 GSM8K, 37.60 MATH) remains very large, highlighting that effective cross-tokenizer knowledge transfer across heterogeneous models is still an open problem.

\begin{table*}[ht]
\centering
\small
\renewcommand{\arraystretch}{1.0}
\begin{tabularx}{0.8\linewidth}{llRR}
\toprule
\textbf{Model} & \textbf{Method} & \textbf{GSM8K} & \textbf{MATH}\\
\midrule\noalign{\vskip -0.6ex}
\rowcolor{gray!20} OpenMath2-Llama3.1-8B & & 87.26 $\color{gray} _{\pm\text{0.92}}$ & 37.60 $\color{gray} _{\pm\text{2.16}}$ \\
\rowcolor{gray!20} Gemma2 2B IT & & 51.48 $\color{gray} _{\pm\text{1.38}}$ & 10.60 $\color{gray} _{\pm\text{1.38}}$ \\
\noalign{\vskip -0.4ex}\midrule
\multirow{5}{*}{Gemma2 2B} & SFT & 59.29 $\color{gray} _{\pm\text{1.35}}$ & 22.40 $\color{gray} _{\pm\text{1.87}}$ \\
& ALM + SFT & 61.56 $\color{gray} _{\pm\text{1.34}}$ & 19.00 $\color{gray} _ {\pm\text{1.76}}$  \\
\cdashline{2-4}\noalign{\vskip 0.5ex}
& Ours & \textbf{62.55} $\color{gray} _{\pm\text{1.33}}$ & 20.08 $\color{gray} _{\pm\text{1.82}}$ \\
\bottomrule
\end{tabularx}
\vspace{-0.3cm}
\caption{Results of cross-tokenizer distilling the large math-specialized OpenMath2-Llama3.1-8B \citep{toshniwal2024openmathinstruct} into the small Gemma2 2B \citep{gemmateam2024gemma2improvingopen} language model. All results are zero-shot CoT.}
\label{table:cross_tokenizer_distil}
\end{table*}

\section{Limitations}
Due to computational constraints, our work explores the task of tokenizer transfer with 3 billion parameter models, and the task of CTD between an 8 billion parameter teacher and a 2 billion parameter student. While these are practical sizes for models that are destined to run on-device, the behavior of CTD methods at larger scales remains underexplored. 

Similarly, our distillation makes use of LORA to reduce the computational requirements, and performing full-parameter optimization may lead to higher performance.

\section{Conclusions}
In this paper we introduced BLD, a simple baseline for cross-tokenizer knowledge distillation that operates through a shared byte-level interface. By converting the teacher's output distribution to byte-level probabilities and attaching a lightweight byte-level decoder head to the student, our method avoids the complex vocabulary alignment procedures required by existing approaches. Despite this simplicity, BLD performs competitively with---and on several benchmarks outperforms---substantially more sophisticated methods across both tokenizer transfer and cross-model distillation settings. The effectiveness of this approach can be enhanced much further, for example, one can use a byte-level transformer architecture as opposed to MLP byte-level heads to capture sequential dependencies at the byte level.

Nevertheless, our experiments reveal a sobering finding: no method, including ours, achieves consistent improvements across all benchmarks and tasks. Performance leadership shifts depending on the benchmark, the transfer target, and the specific model pair. This inconsistency suggests that cross-tokenizer distillation remains a fundamentally open problem. We thus encourage the community to continue pursuing this line of research which has strong practical implications.

\bibliography{custom}

\appendix

\section{Evaluating the use of Linear Layers as Byte Level Heads}
\label{app:byte_level_head}
\begin{figure*}[t]
\centering
\includegraphics[width=0.49\linewidth]{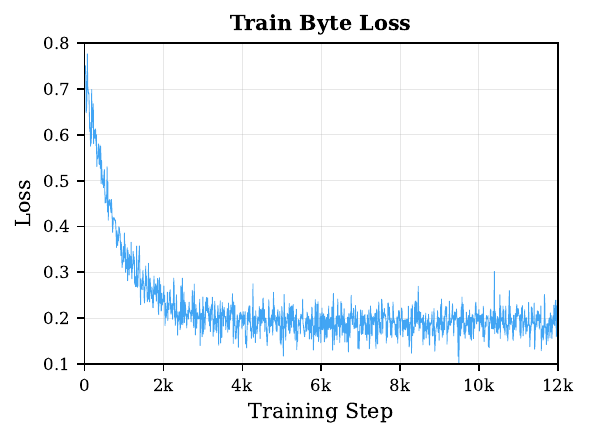}
\includegraphics[width=0.49\linewidth]{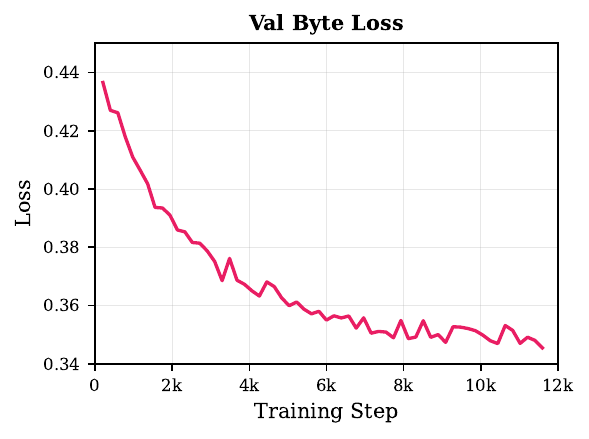}
\includegraphics[width=0.49\linewidth]{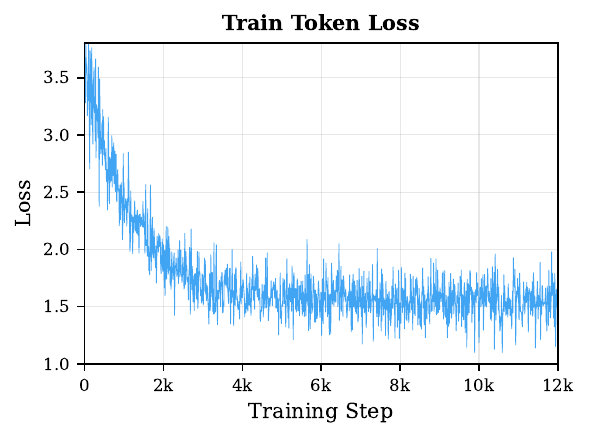}
\includegraphics[width=0.49\linewidth]{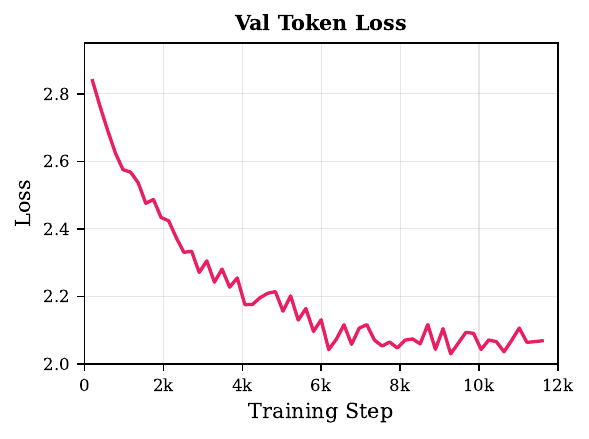}
\caption{Training and validation losses for a Llama3.2-1B model with added byte-level head trained on a subset of the TULU-3 dataset with supervised fine-tuning only at the byte level. The top row plots are for training curves, while bottom row ones are for validation.}
\label{fig:byte_sft}
\end{figure*}

To test the effectiveness of a simple linear layer for each byte level head, we performed SFT \textit{only at the byte level} on the Llama3.2 1B model \citep{grattafiori2024llama3herdmodels} on a subset the TULU-3 dataset \citep{lambert2024tulu3}. We then looked at training and validation losses over both bytes and tokens. We report the plots in Figure \ref{fig:byte_sft}. We observe that, not only do the training and validation losses decrease smoothly for the byte level, but, surprisingly, they decrease also for the token level, demonstrating the effectiveness of adding even simple linear layers as heads for the byte level interface. This also indicates that a byte-level probability distribution can be effectively used for knowledge distillation -- thus bridging a gap between different tokenizers with a common byte-level interface. 

\section{Training Hyperparameters}    
\label{app:hyperparameters}
We provide the values for the main hyperparameters used in our experiments, together with the respective search space for the tuning procedure in Table \ref{tab:hyperparameters}. The values for the baselines follow the optimized setup of \citet{Minixhofer2025}, and only the learning rate has been further tuned due to computational constraints. For our method we tested different values of the weights for the loss functions.

\begin{table*}[h]
    \centering
   \resizebox{\textwidth}{!}{%
    \begin{tabular}{lp{4.5cm}l}
    \toprule 
    \textbf{Hyperparameter} & \textbf{Value} & \textbf{Search Space} \\
    \midrule 
    \multicolumn{3}{l}{\textit{LoRA}} \\
        \quad Rank ($r$) & 64 & --- \\ 
        \quad Alpha ($\alpha$) & 64 & --- \\
        \quad Dropout & 0.05 & --- \\
        \quad Target modules & \texttt{q\_proj}, \texttt{v\_proj}, \texttt{k\_proj}, \texttt{o\_proj}, \texttt{gate\_proj}, \texttt{up\_proj}, \texttt{down\_proj} \\
    \midrule 
    \multicolumn{3}{l}{\textit{Optimiser}} \\
        \quad Algorithm & AdamW & --- \\ 
        \quad Learning rate & $2 \times 10^{-5}$ & $\{5\text{e-}6,\ 1\text{e-}5,\ 2\text{e-}5,\ 3\text{e-}5,\ 5\text{e-}5,\ 1\text{e-}4\}$ \\
        \quad Weight decay & 0.01 & --- \\ 
        \quad $(\beta_1, \beta_2)$ & $(0.9,\; 0.95)$ & --- \\
        \quad Gradient clipping (norm) & 1.0 & --- \\
    \midrule 
    \multicolumn{3}{l}{\textit{Learning rate schedule}} \\ 
        \quad Scheduler & Cosine + linear warm-up & --- \\
        \quad Warm-up steps & 1{,}000 & --- \\ 
    \midrule
    \multicolumn{3}{l}{\textit{Training}} \\ 
        \quad Epochs & 5 & --- \\
        \quad Batch size (per device) & 2 & --- \\ 
        \quad Gradient accumulation steps & 4 & --- \\
        \quad Max sequence length & 512 & --- \\ 
        \quad Precision & \texttt{bf16-mixed} & --- \\
    \midrule 
    \multicolumn{3}{l}{\textit{Loss coefficients}} \\
        \quad KL divergence ($\lambda_{\mathrm{KL}}$) & 0.1 & $\{0.1,\ 0.2,\ 0.5,\ 0.8,\ 1.0\}$ \\ 
        \quad Byte SFT ($\lambda_{b}$) & 1.0 & $\{0.5,\ 1.0\}$ \\
    \midrule 
    \multicolumn{3}{l}{\textit{Byte-level decoder head}} \\
        \quad Parallel heads & 10 & --- \\ 
        \quad Byte vocabulary size & 261 & --- \\
    \bottomrule
    \end{tabular}
    }
    \caption{Training hyperparameters used in all experiments. The \textit{Search Space} column lists the values explored during hyperparameter tuning; a dash indicates the value was fixed without search.}
    \label{tab:hyperparameters}
\end{table*}                                                                                                          
                                                     
\section{Approximation Settings for Byte-Probability Computations}
\label{app:approx_settings}

The algorithm proposed by \citet{vieira2025from} provides an efficient approximation for computing byte-level probabilities from a token-level language model. In this section we describe the approximation parameters used in our implementation and the empirical procedure used to select them.

\subsection{Approximation Parameters}

The algorithm introduces two parameters, $K$ and $\epsilon$, that control the trade-off between computational efficiency and approximation accuracy when estimating the byte-level probability

\[
P_T(b_1,b_2,\dots,b_{N_b})
\]

from a teacher model $f_t$ operating over a token vocabulary (see Section~\ref{sec:preliminaries}).

\paragraph{Beam width ($K$).}
The algorithm performs a beam search over token sequences that are compatible with a given byte prefix.
The beam width $K$ specifies the maximum number of hypotheses retained during the search. Larger values of $K$ allow more tokenization paths to be explored, which improves approximation accuracy but increases computational cost.

\paragraph{Pruning threshold ($\epsilon$).}
During beam search, hypotheses with very small probability mass are removed.
Specifically, beams whose probability falls below a threshold $\epsilon$ relative to the highest-probability beam are pruned.
This pruning step eliminates tokenization paths that contribute negligibly to the final byte probability distribution.

Together, $K$ and $\epsilon$ determine the number of tokenization paths considered during the computation.

\subsection{Algorithm for Byte Probability Computation}

The byte-level probability distribution at each position is computed using the following procedure:

\begin{enumerate}
\item \textbf{Initialization:} Create a beam state with parameters $K$ (beam width) and $\epsilon$ (pruning threshold). The beam maintains a set of candidate tokenization paths, each with an associated probability weight.

\item \textbf{For each byte position $i$:}
\begin{itemize}
    \item \textbf{Compute distribution:} Call \texttt{logp\_next()} to obtain the log probability distribution over the next 256 possible byte values. This operation marginalizes over all tokenization paths in the current beam:
    \begin{align}
    \log P(b_i & \mid b_{<i}) = \nonumber \\ 
    &\log \sum_{t \in \text{Beam}} P(t) \cdot P(b_i \mid t)
    \end{align}
    where $t$ represents a tokenization path and $P(t)$ is its weight.

    \item \textbf{Advance beam:} Incorporate the observed byte $b_i$ into the beam using the operation \texttt{beam.prune() << byte}. This extends each candidate path by consuming the byte.

    \item \textbf{Prune paths:} Remove tokenization paths with probability below the threshold $\epsilon$ relative to the highest-probability path. Retain at most $K$ paths.

    \item \textbf{Handle token boundaries:} When a path completes a token, extend the beam by starting a new token using the teacher model's next-token probabilities.
\end{itemize}

\end{enumerate}

The key computational bottleneck is the teacher model inference at token boundaries. The beam parameters $K$ and $\epsilon$ control how many tokenization alternatives are maintained, which determines both accuracy and computational cost.

\begin{figure*}[t]
\centering
\begin{minipage}[t]{0.48\textwidth}
\centering
\includegraphics[width=\linewidth]{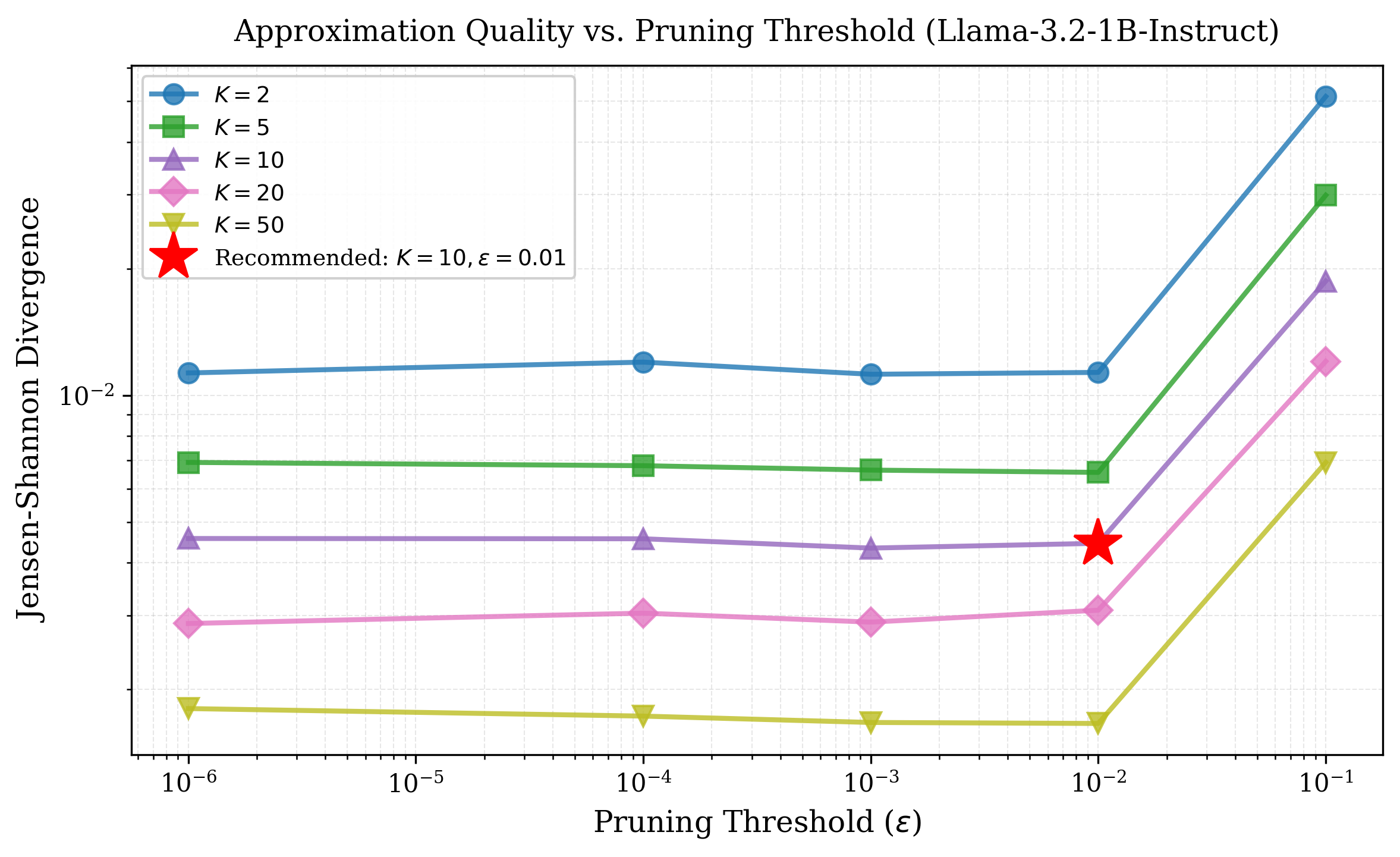}
\caption{Jensen–Shannon divergence between approximated byte distributions and the reference distribution under different approximation settings.}
\label{fig:jsd}
\end{minipage}
\hfill
\begin{minipage}[t]{0.48\textwidth}
\centering
\includegraphics[width=\linewidth]{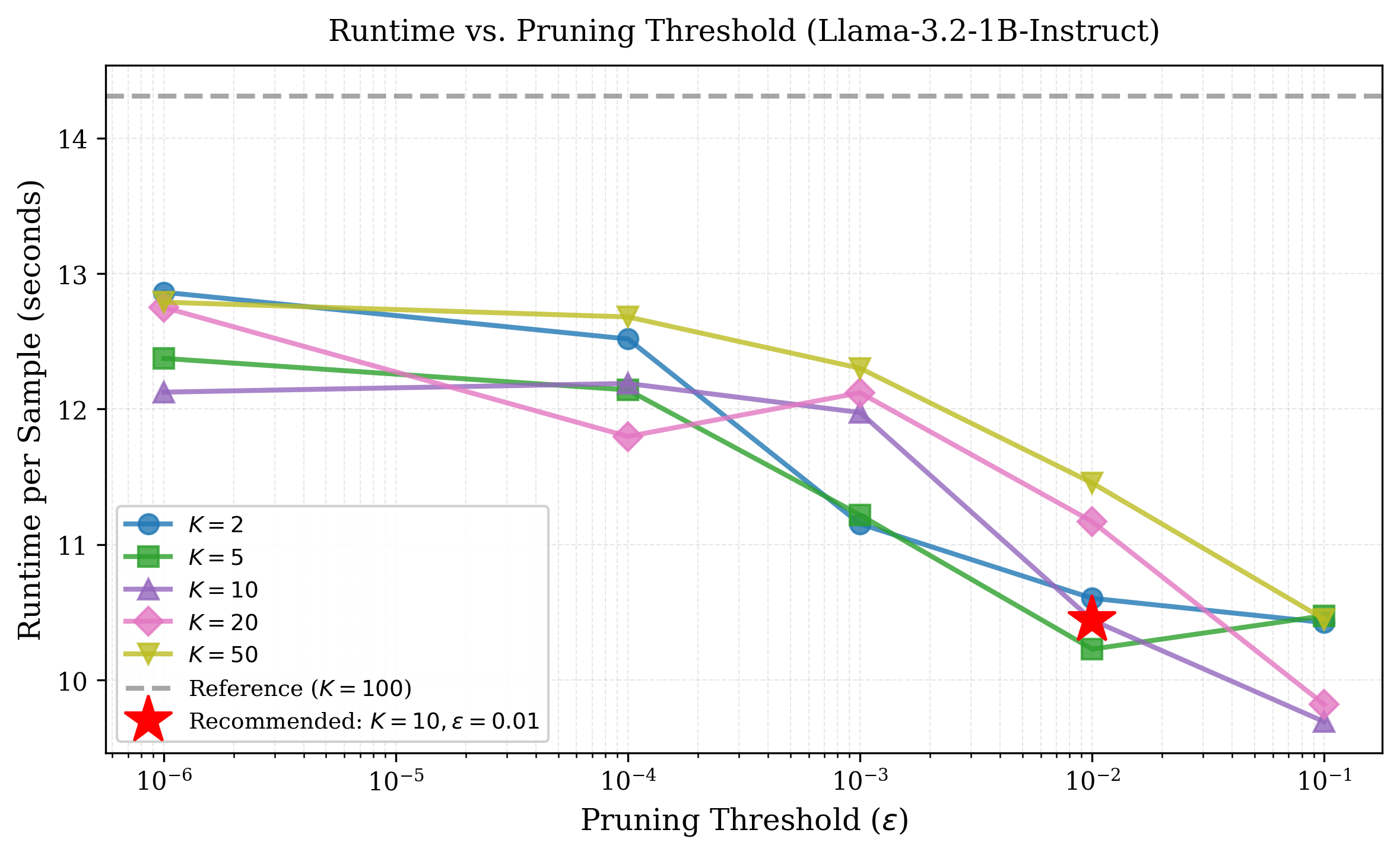}
\caption{Runtime of byte-probability computation under different beam search configurations.}
\label{fig:runtime}
\end{minipage}
\end{figure*}

\subsection{Evaluating Approximation Quality}

We measure the Jensen–Shannon divergence (JSD) between the approximated byte probability distribution and a high-precision reference distribution computed using $K = 100, \epsilon = 10^{-6}$.
Figure~\ref{fig:jsd} shows the resulting approximation error for different combinations of $K$ and $\epsilon$.
We observe that the setting $K = 10, \epsilon = 0.01$ achieves a Jensen--Shannon divergence of 0.0045. Figure~\ref{fig:runtime} shows that runtime is primarily affected by the pruning threshold $\epsilon$ (lower values retain more beams), while beam width $K$ has minimal impact due to efficient GPU batching of token queries. We also evaluate the effect of the approximation on downstream distillation performance by measuring the distilled model's perplexity and task accuracy, confirming that configurations with JSD $< 0.005$ produce negligible performance degradation.

\subsection{Experimental Setup}

We conduct experiments using Llama-3.2-1B-Instruct and Llama-3.2-3B-Instruct as teacher models, with the Tulu-3 dataset for distillation. We test beam widths $K \in \{2, 5, 10, 20, 50, 100\}$ and pruning thresholds $\epsilon \in \{10^{-1}, 10^{-2}, 10^{-3}, 10^{-4}, 10^{-6}\}$, measuring runtime and JSD relative to the reference configuration for each setting.

\subsection{Parallel Implementation}

Our implementation achieves efficient throughput through multi-level parallelization. The dataset is partitioned into shards, with each shard processed by an independent worker (one per GPU). Within each worker, we use a process pool with \texttt{n\_sample\_worker=15} to parallelize across samples, and the underlying trie operations batch up to 1000 token probability queries per forward pass. We use Python's \texttt{asyncio} framework to overlap CPU preprocessing with GPU computation.

In our experiments using four NVIDIA RTX 3090 GPUs, the configuration $K=10, \epsilon=0.01$ achieves approximately 10.4 seconds per sample for 100--150 byte sequences. We choose this configuration because it provides excellent approximation accuracy (JSD $< 0.005$) while using 10$\times$ less memory than the reference configuration ($K=100$), enabling higher sample-level parallelism. The lower memory footprint allows us to process more samples concurrently, and the balanced pruning threshold $\epsilon=0.01$ avoids both overly aggressive pruning (which degrades accuracy) and overly conservative retention (which increases memory usage). With this configuration and parallelism, computing byte probabilities for the entire Tulu-3 dataset requires approximately 2 days.

\end{document}